\documentclass[11pt]{amsart}
\usepackage{geometry}                
\usepackage{graphicx}
\usepackage{amssymb}
\usepackage{epstopdf}
\usepackage{url}
\usepackage{array}
\usepackage{paralist} 
\usepackage{verbatim} 
\usepackage{subfig}

\title{Benchmarking Named Entity Disambiguation approaches for Streaming Graphs}
\author{Sutanay Choudhury, Chase Dowling}

\begin{document}
\maketitle

\section{Introduction}

Named Entity Disambiaguation (NED) is a central task for applications dealing with natural language text.  Assume that we have a graph based knowledge base (subsequently referred as Knowledge Graph) where nodes represent various real world entities such as people, location, organization and concepts.  Given data sources such as social media streams and web pages Entity Linking is the task of mapping named entities that are extracted from the data to those present in the Knowledge Graph.  This is an inherently difficult task due to several reasons.  Almost all these data sources are generated without any formal ontology;  the unstructured nature of the input, limited context and the ambiguity involved when multiple entities are mapped to the same name make this a hard task.

Our exploration is motivated by the objective of incremental maintenance of Knowledge Graphs, where we continuously monitor data streams and update the Knowledge Graph with information from newer events.  A good Entity Linking/Named Entity Disambiguation solution needs to perform well on three aspects for our target setting.

\begin{enumerate}
\item \textsc{Scalability}  The solution needs to support very large Knowledge Graphs.  As the graphs grow in size, the number of candidates for mapping an entity increases.    
\item \textsc{Speed}  Throughput is an important factor in a streaming setting.  We can not afford to select algorithms that do not satisfy a runtime constraint.
\item \textsc{Accuracy}  For the same algorithm, its accuracy can vary depending on the nature of the input.  The same entity linking system can perform poorly as we go from well structured wikipedia pages to short text excerpts from social media.
\end{enumerate}

With the above in mind, we looked at two state of the art systems employing two distinctive approaches: graph based Accurate Online Disambiguation of Entities (AIDA) and Mined Evidence Named Entity Disambiguation (MENED), which employs a statistical inference approach.  We compare both approaches using the data set and queries provided by the Knowledge Base Population (KBP) track at 2011 NIST Text Analytics Conference (TAC).  This report begins with an overview of the respective approaches, followed by detailed description of the experimental setup.  It concludes with our findings from the benchmarking exercise.  

\section{Background}

Within in the broader field of named entity disambiguation, we focus on Knowledge Base Population problem as introduced by the NIST Text Analytics Conference (TAC) \cite{ji2011knowledge}.  This competition uses a 2008 version of Wikipedia as the Knowledge Base (KB).  Given a string referring to a person, organization or geo-political entity name and a background document providing contextual information, the entity linking task requires a system to return the ID of an entity in the knowledge base if it exists, else return NIL.  

\subsection{Problem Statement}

We have a knowledge base $KB = (D, F, N)$ where $D$, $F$ and $N$ represent a collection of documents, features and names. $C$ contains information about a set of entities $E$. Each $d_k \in D$ corresponds to an entity $e_i \in E$, and has a feature vector $F_k$ and a name $N_k$ associated with it. A query is a three-tuple $Q = (Q_t, Q_n, Q_d)$. $Q_n$ indicates the name of the queried entity. $Q_t$ is the type of the queried entity, which is one of the following: (a) person, (b) geo-political entity, (c) topic and (d) unknown. The query contains a sample document describing the query context, which is referred to as $Q_d$.   As an example, a query may contain ``John" as the query name, "person" as query type and a sample publication to provide the contextual information.

\textsc{Problem Definition} Given a query $Q$, the goal of the entity disambiguation system is to provide a query function $f(Q)$ that returns the most relevant entity in the knowledge base if it exists, or NIL if it does not exist. \\

The entity linking process contains 3 steps: 1) pre-processing the query to clean or expand into multiple forms using domain background 2) generating a list of all candidate entities in the KB that are potential matches for the query and 3) ranking the candidates in terms of match quality.    

\subsection{Graph-based approach} 

Given a query string name and a background document, a graph based approach will extracts all potential entities from the background document.  It builds a query graph that includes the entity name, and other potential entities connected via appropriate relations.  Given this query graph, the algorithm selects the entity in the Knowledge Graph whose neighborhood provides closest mapping to the query graph.  Finding the distances between two entities is a critical step in the computation process for entity disambiguation solutions that rely on a graph based approach.  The distance computation can be performed by considering multiple factors such as semantic association, specialization of the semantic association, association length and rarity of associations \cite{aleman2005ranking, fang2011rex}.  

We evaluated the AIDA algorithm developed by Hoffart et al. \cite{hoffart2011robust}.  AIDA identifies all meaningful entities within a document (ignoring personal pronouns, etc) and assigns them a label corresponding to a document within the Knowledge Graph.  By considering multiple entities within a document simultaneously, the algorithm boils down to a graph matching problem finding expected concurrent entities within the Knowledge Graph.  Model parameters are encoded in entity linking priors (probability that any two Knowledge Graph entities are semantically related) and remained unchanged within the Knowledge Graph.  AIDA requires a trained and annotated graphical version of Wikipedia to serve as a Knowledge Graph, however, no other preprocessing is required;  

\subsection{Statistical approach}  
This class of solutions exploit that fact that the context for any entity (query or candidate) can be distinguished by a set of representative words.  The reason why context can help disambiguation is that each referent entity candidate can be distinguished by a set of representative words. Those representative words can be seen as the disambiguation evidences for those entity candidates. Therefore it is natural to model each entity as a topic and imagine those representative words are generated from such topics.  Thus a representative document for an entity such as a Wikipedia article can be viewed as an unique topic composed of the words.  Given a query entity such algorithms will choose from candidates that is most topically coherent.    

We implemented the MENED algorithm developed by Li et al. \cite{li2013mining}.  MENED is a semi-supervised learning algorithm employing a generative version of Latent Dirichlet Allocation \cite{blei2003latent}.  MENED was designed for short text queries, using word concurrence to search for candidate document labels before performing a generative version of LDA on all candidate document labels to infer the correct label for the query document.

\section{Comparison of disambiguation algorithms}

\textsc{Configuring AIDA} AIDA is available via Github and version stability is ensured by a web-based build/test server. The AIDA graph matching algorithm and dependencies (e.g. Stanford NER) are combined into a single Java package and can be run as a restful API.  Because learned parameters are encoded in AIDA's knowledge base, multiple instances of AIDA can be run concurrently. 

\textsc{Configuring MENED} Parameter and hyper-parameter selection is driven by small samples of the query corpus.  Publication parameters ($\alpha = 0.001$, $\alpha_{df} = 0.01$, $\beta = 0.001$, $\beta_{df} = 0.01$, $\beta_{bg} = 0.1$, $\gamma_{1} = 0.0003$, $\gamma_{2} = 0.001$) were accurate to the order of magnitude after sampling our query data set and were left unchanged.  Unprocessed TAC 2009 Wikipedia text data was used as the knowledge base for experimental reproduction of the authors' results.  Data was stored in an Elastic Search database, a restful API serving JSON objects built on Apache Lucene for cloud databasing applications.  Search results are scored using Lucene's built-in TF-IDF relevancy scoring.  Query data and mined candidate entity data were preprocessed by removing stop words and roughly tokenized (removing punctuation and attempting to capture unrelated text such as URL's).

\subsection{Software Usability/Scaling}

\begin{figure}[htbp]
  \centering
  \includegraphics[scale=0.5]{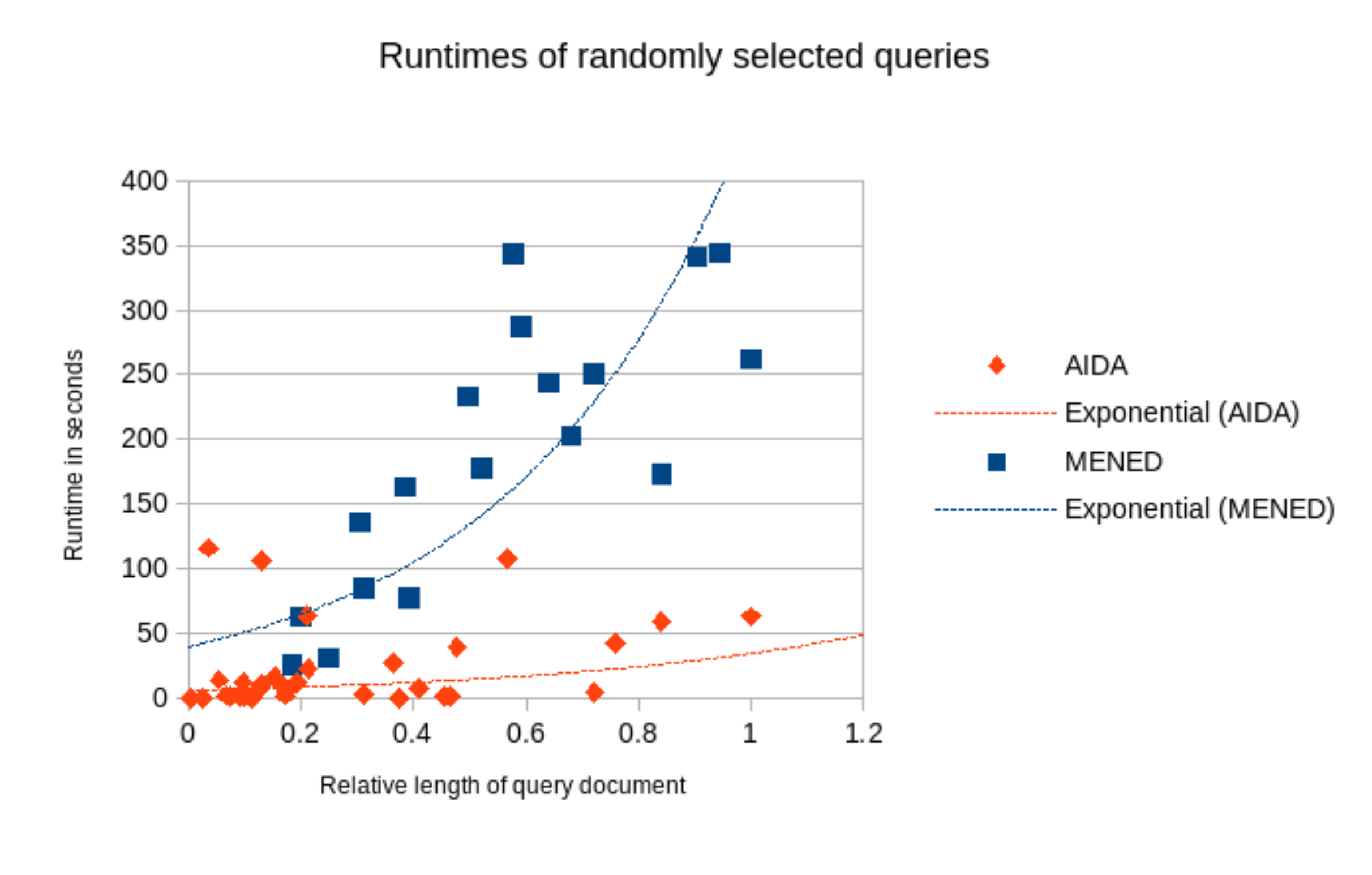}
  \caption{TAC Benchmark runtimes for AIDA and MENED}
  \label{fig:tac_benchmark_runtimes}
\end{figure}

Figure \ref{fig:tac_benchmark_runtimes} shows the runtimes for both AIDA and MENED running the TAC benchmark.  AIDA's runtime appears to increase linearly as a function of entity mentions and MENED's appears to increase linearly as a function of document length although we suspect polynomial complexity.  Classical LDA is known to be NP-Hard \cite{sontag2011complexity}.  Neither publication provides explicit runtime complexities, however, AIDA claims to be designed for online applications and empirical measurements of runtime in seconds showed most disambiguation tasks taking less than 60 seconds.  We restricted maximum runtimes to 600 seconds.

Nearly all of AIDA's incorrect label assignments came from forcing a label on a ÒNILÓ or wildcard query document based on the TAC benchmark's ground truth.  Although AIDA should exceed 80\% for general queries, we were able to achieve ~79\% accuracy on TAC benchmark data.  TAC benchmark data assigns a single label to entire documents, while AIDA identifies labels for all named entities within a document.  As long as one of the disambiguated entities from within the document corresponded with the document label, we counted the disambiguation as correct.  We did not attempt to set a minimum threshold of certainty for which the ÒNILÓ label would be selected as query document lengths varied greatly.

We were unable to achieve MENED's claimed accuracy of approximately 70-80\% for two reasons: 1) the MENED algorithm used the Google Search API combined with a Wikipedia repository of 2012 as a knowledge base whereas we made use of the TAC 2009 Wiki raw text dump only and 2) we were unable to specify the conditions under which query documents were curated in the original test (e.g. when original authors tested MENED on Twitter posts, tweets were hand selected for disambiguation in order to create a set with ground truth available).  At best we achieve approximately 35\% accuracy, correctly assigning ÒNILÓ to all wildcard query documents and in disambiguating hyper-specific queries.

In making use of the Google Search API, MENED acquires an implicit measure of search relevance in entity candidate selection (e.g. searching ÒQueen ElizabethÓ invariably yields the current monarch as opposed to her predecessor).  Moreover, searches of our knowledge base were constrained to TF-IDF measures of document relevancy.  ÒQueen ElizabethÓ happens to have the greatest weight in the Wikipedia article about the monarchy of New Zealand, where the extremely short summary mentions Queen Elizabeth II three times.

\subsection{Example queries}

We produce two example queries:

1: "Einstein was born in Ulm."\\

2: "When Page played Kashmir at Knebworth, his Les Paul was uniquely tuned."\\

AIDA produces the following annotated text for 1: \url{http://en.wikipedia.org/wiki/Albert_Einstein} Einstein was born in \url{http://en.wikipedia.org/wiki/Ulm} Ulm. and for 2: When Page played \url{http://en.wikipedia.org/wiki/Kashmir_(song)} Kashmir at \url{http://en.wikipedia.org/wiki/Knebworth_Festival_1979} Knebworth, his \url{http://en.wikipedia.org/wiki/Gibson_Les_Paul} Les Paul was uniquely tuned.


MENED incorrectly labels the first sentence with "NIL" as the most likely label, following by "Mileva Mari", a fellow student and Einsteins first wife.  The second setence is labeled with the top three results being "Les\_pauvres\_riches", "Les\_Contrebandi", and "Kashmir" the region.  This largely due to conflicting entity concurrence in the query and entity search relavance ("Kashmir" the song, when searched directly using TF-IDF and without the Google Search API, is the 5th most likely result).

\section{Conclusions and Future Research}

We compare two, reportedly robust entity disambiguation algorithms using the 2011 TAC entity disambiguation benchmark dataset.  Each system was evaluated with a Cloud-based deployment in mind.  The MENED algorithm was implemented using a Cloud based database, while the AIDA algorithm was deployed and used as a web service.  Although the goal of each algorithm differs slightly, we achieve high accuracy with AIDA, and moderate to low accuracy using MENED.  Although MENED attempts to apply a more flexible label to an entire document, AIDA treats individually identified entities within a document with high accuracy.  We recommend AIDA for anyone seeking an out of the box solution.  Its high accuracy is also a testament to the relevance of graph based approaches in Natural Language Processing.  \\

One of MENED's novel contributions is that with streaming data such as social media text, it updates the knowledge base with newly inferred facts to improve query processing accuracy.  This ostensibly allows the knowledge base to handle more vague, up to the minute queries and attribute black swan type events.  We propose integrating MENED's machine learning approaches with AIDA's graph search based approach as a future research path.  This will enable updating the Knowledge Graph by adding new semantic relationships, allowing stale relationships to fall off, and updating relationship probability priors as inferences are made on streaming data.     

\bibliographystyle{abbrv}
\bibliography{cassnlp}

\begin{thebibliography}{1}

\bibitem{aleman2005ranking}
B.~Aleman-Meza, C.~Halaschek-Weiner, I.~B. Arpinar, C.~Ramakrishnan, and A.~P.
  Sheth.
\newblock Ranking complex relationships on the semantic web.
\newblock {\em Internet Computing, IEEE}, 9(3):37--44, 2005.

\bibitem{blei2003latent}
D.~M. Blei, A.~Y. Ng, and M.~I. Jordan.
\newblock Latent dirichlet allocation.
\newblock {\em the Journal of machine Learning research}, 3:993--1022, 2003.

\bibitem{fang2011rex}
L.~Fang, A.~D. Sarma, C.~Yu, and P.~Bohannon.
\newblock Rex: Explaining relationships between entity pairs.
\newblock {\em Proceedings of the VLDB Endowment}, 5(3):241--252, 2011.

\bibitem{hoffart2011robust}
J.~Hoffart, M.~A. Yosef, I.~Bordino, H.~F{\"u}rstenau, M.~Pinkal, M.~Spaniol,
  B.~Taneva, S.~Thater, and G.~Weikum.
\newblock Robust disambiguation of named entities in text.
\newblock In {\em Proceedings of the Conference on Empirical Methods in Natural
  Language Processing}, pages 782--792. Association for Computational
  Linguistics, 2011.

\bibitem{ji2011knowledge}
H.~Ji and R.~Grishman.
\newblock Knowledge base population: Successful approaches and challenges.
\newblock In {\em Proceedings of the 49th Annual Meeting of the Association for
  Computational Linguistics: Human Language Technologies-Volume 1}, pages
  1148--1158. Association for Computational Linguistics, 2011.

\bibitem{li2013mining}
Y.~Li, C.~Wang, F.~Han, J.~Han, D.~Roth, and X.~Yan.
\newblock Mining evidences for named entity disambiguation.
\newblock In {\em Proceedings of the 19th ACM SIGKDD international conference
  on Knowledge discovery and data mining}, pages 1070--1078. ACM, 2013.

\bibitem{sontag2011complexity}
D.~Sontag and D.~Roy.
\newblock Complexity of inference in latent dirichlet allocation.
\newblock In {\em NIPS}, pages 1008--1016, 2011.

\end{thebibliography}

\end{document}